# Clustering in Recurrent Neural Networks for Micro-Segmentation using Spending Personality

Charl Maree[1]
Center for AI Research
University of Agder
Grimstad, Norway
charl.maree@sr-bank.no

Christian W. Omlin
Center for AI Research
University of Agder
Grimstad, Norway
christian.omlin@uia.no

*Abstract*—Customer segmentation has long been a productive field in banking. However, with new approaches to traditional problems come new opportunities. Fine-grained customer segments are notoriously elusive and one method of obtaining them is through feature extraction. It is possible to assign coefficients of standard personality traits to financial transaction classes aggregated over time. However, we have found that the clusters formed are not sufficiently discriminatory for micro-segmentation. In a novel approach, we extract temporal features with continuous values from the hidden states of neural networks predicting customers' spending personality from their financial transactions. We consider both temporal and non-sequential models, using long short-term memory (LSTM) and feed-forward neural networks, respectively. We found that recurrent neural networks produce micro-segments where feed-forward networks produce only coarse segments. Finally, we show that classification using these extracted features performs at least as well as bespoke models on two common metrics, namely loan default rate and customer liquidity index.

## I. INTRODUCTION

Effective customer engagement is critical in any retail industry and retail banking is no exception. As their customer bases grow, banks have to employ ever advancing tools to maintain if not improve the level of personalization in their interactions. AI provides such a tool and is becoming ubiquitous in retail banking [1]. In machine learning, feature extraction is the process of compressing information held in a feature set and replicating it with high fidelity using fewer features. In contrast to feature selection which selects a subset of features, feature extraction creates new features with reduced redundancy. A feature is a single quantifiable property of the data and can be numerical, categorical or textual. Dimensionality reduction is important in applications where independent data observations are finite; an increasing number of features rapidly increases the volume of the feature space such that available data quickly become sparse, counteracting the statistical significance of results. Feature extraction may also facilitate the prediction of a different but related dataset, e.g., through transfer learning which applies previously learned knowledge to a new problem. The learned relationships between the original features and the reduced features may be retained when predicting new dependent variables with fewer observations. The success of transfer learning has, for example, been demonstrated for image classification [2]. By re-using these pretrained models, smaller teams may benefit from their exceptional properties while forgoing the majority of data acquisition and preprocessing.

Transfer learning has been used in banking related applications such as customer credit scoring where knowledge was transferred across different geographical districts, and customer churn prediction where knowledge was transferred across different time periods and districts [3] [4].

Customer micro-segmentation is a promising application of AI in banking; in order to develop personalized products and services, it is important to differentiate between different types of customers [5]. Traditional customer segmentation classifies individuals along demographics such as age, gender, location, etc. so as to optimize customer interactions [6]. It produces a coarse classification which could fail to depict nuanced differences between individuals, potentially leading to discrimination e.g., in credit rating according to postal codes [7]. In contrast, micro-segmentation provides a more sophisticated classification of customers and therefore holds immense potential for personalized financial products and services. Despite the advantages, there have been no published applications of micro-segmentation of financial customers. We provide a solution through a novel approach in which we extract temporal features from the states of a recurrent neural network. We show that these features form hierarchical clusters that facilitate micro-segmentation.

We intend to develop personalized digital financial advisors that match individual customers' personalities. There is a documented correlation between financial transactions and personality [8] and evidence that spending according to personality increases happiness [9]. In this study, we extract features from the financial transactions of ca. 26,000 customers over six years. We compare the performance of feed-forward neural networks to that of recurrent neural networks in micro-segmentation; to the best of our knowledge, explicit temporal modelling of customer spending behavior has never been considered before. We show that in the state space, customer spending follows 'ski slopes', i.e., well-defined discrete trajectories with a low average change of direction. In addition, these trajectories cluster for both dominant and lesser personality traits. These trajectories are promising salient features that are novel and have the potential to be used as the basis for future personalized financial products and services.

Finally, we demonstrate the efficacy of the extracted features in a transfer learning case study predicting two common customer metrics, namely loan default rate and customer liquidity index; we show that the extracted features performed at least as well as randomly initialized models trained on larger datasets. Using these extracted features, we intend to perform a micro-segmentation of our customers to facilitate the development of personalized financial advisors.

[1]Strategy Innovation and Development, SpareBank 1 SR-Bank ASA, Norway.
XXX-X-XXXX-XXXX-X/XX/$XX.00 ©20XX IEEE

This research was partially funded by a grant from The Norwegian Research Council; project nr 311465.

## II. Related Work

Spending as evinced in financial transactions has been proven to be a promising personality predictor. In [8] the authors used a random forest to predict the Big Five personality traits – openness, conscientiousness, extraversion, agreeableness, neuroticism – from the transactions of 2,193 banking customers. They determined customer personality through the Big-Five Inventory-10 questionnaire [10]. Their reported accuracy was comparable to that of using demographics as a predictor, but they reported a higher accuracy when using more specific personality traits such as materialism and self-control. An earlier paper by the same authors also used the Big Five model and a questionnaire to determine customers' personalities [9]. They then derived a set of coefficients – between -3 and 3 – associating 59 transaction classes with each of the Big Five personality traits. A panel of 100 evaluators rated each class' correlation with each of the Big Five traits, from which they determined a mean correlation. An example from their study is a coefficient of -0.82 for the trait "extraversion" and the spending category "books" which suggested a mild negative association between buying books and extraversion. They used these coefficients to investigate the relationship between customer spending and their personalities and reported a causal relationship between personality-oriented spending and happiness; such spending outweighed the effect of total income. Two independent studies also used the Big Five model and found correlations between personality traits and spending [11] [12]. There clearly exists a correlation between consumer spending and personality, and the Big Five model has been a popular model for personality classification.

Generally, there are surprisingly few publications on micro-segmentation and none in the field of finance and banking. One notable publication achieved a coarse segmentation through feature extraction using customers' Big Five personality traits along with traditional demographics and transactional data [13]. They trained both an unsupervised autoencoder and a supervised neural network with loan default probability as output. The goal was to extract features that analysts may easily visualize. They concluded that by including personality, the prediction accuracy of loan defaults improved, and they showed that they were able to cluster customers in a low dimensional space.

## III. Empirical Methodology

We extracted features from customer spending data using feed-forward and recurrent neural networks with both unsupervised (autoencoder) and supervised (predictor) architectures. We then investigated the efficacy of the extracted features in a transfer learning case study predicting loan default rate and customer liquidity index.

### A. Data

For our dataset we used the financial transactions of ca. 26,000 anonymous customers between the ages of 30 and 60 over a period of 6 years. The transactions were classified into categories, such as "groceries", "transportation", "savings", etc. using the explainable AI system detailed in [14]. We then added an element of time by aggregating the transactions of each customer annually and by transaction category, normalized by annual income; each datapoint represented the annual spending distribution of each customer across the transaction categories in six time-steps. We formatted the dataset to support two types of neural networks: feed-forward and recurrent. The dataset for the feed-forward network had the shape $[n \times 6, m]$ where $n = 26,000$ customers and $m$ is the number of transaction categories. In the dataset for the recurrent network, each customer had a sequence of 6 time-steps resulting in the data shape $[n, 6, m]$. We split the data into training (80%) and validation (20%) sets and ran 20 experiments with randomly sampled data from the training set to determine the accuracy and confidence intervals.

### B. Spending-Evinced Personality

We used the coefficients published in [9] to calculate the Big Five personality traits from our aggregated transactions. For each customer, we calculated both annual and overall personality types across the 6-year period resulting in two datasets: a $[n \times 6, 1]$ and a $[n, 1]$ dataset respectively. A customer's dominant personality trait is a delicate concept and one that is useful to introduce; we defined it as the personality trait that, of the five, had the highest absolute value. For example, a large negative extraversion score translates to a large positive introversion score; in comparisons between the traits, the absolute value must therefore be used.

### C. Feature Extraction

Feature extraction and dimensionality reduction is a mainstay in machine learning. A widely used method is principal component analysis [15]. It only identifies linear correlations between features, a shortcoming addressed by autoencoders [16]. An autoencoder is an unsupervised neural network that aims at reconstructing input data in the output layer [17]. The information is compressed by successively reducing the number of nodes in the hidden layers to reach a bottleneck. It thus learns a feature representation for a set of data with a reduced dimensionality. The underlying assumption is that these features are salient since they are able to reconstruct the information contained in the input data.

We used four neural network architectures to extract features from our classified transactions: a *feed-forward autoencoder* accepting $[n \times 6, m]$ customer spending observations as both input and output, a *feed-forward predictor* with the same input but $[n \times 6, 1]$ annual personality traits as output, a *recurrent autoencoder* accepting $[n, 6, m]$ sequential spending observations as both input and output, and a *recurrent predictor* with the same input but $[n, 1]$ personality types as output. The recurrent networks used long short-term memory (LSTM) nodes, which has been described in, e.g., [18]. The size of the networks (number of nodes and layers) were hyperparameters and optimized for each network architecture.

### D. Transfer Learning

Transfer learning – for which most of the weights of a neural network are pre-trained on a related supervised machine learning task – significantly reduces the number of samples needed in training. Knowledge may also be extracted from recurrent neural networks, as demonstrated in [19]. In this early work, the authors investigated the internal neuron activations of recurrent neural networks and managed to extract the rules that govern the model. The same authors in [20] were some of the first to demonstrate transfer learning in recurrent neural networks by initializing the network with weights learned on another dataset.

We compared the performance of the extracted features from the predictive feed-forward and recurrent neural networks to that of randomly initialized networks of identical architectures. In this case study, we predicted two common metrics in banking: loan default rate and customer liquidity index. For the transfer learning models, we initialized the weights with those from our pretrained models, while the baseline models were randomly initialized. Pretrained weights were non-trainable. For training, we used a reduced training set of 100 randomly selected observations and ran 20 experiments to calculate the accuracy and confidence intervals. Accuracy was measured against a large validation set of ca. 5,000 observations.

## IV. Results and Discussion

### A. Micro-Segmentation through Feature Extraction

Firstly, we found that the raw personality data – the Big Five personality scores – naturally formed fuzzy clusters along the most dominant personality trait and along that specific axis. We illustrate this phenomenon in Fig. 1 where all the points to the right of a given threshold – the vertical dotted line – represent individuals whose dominant personality trait is 'openness'. These points naturally form a fuzzy cluster to the right of this threshold. However, we observed inconsistent customer spending patterns for shorter time windows leading to unstable clusters with customers appearing in different clusters for different time windows.

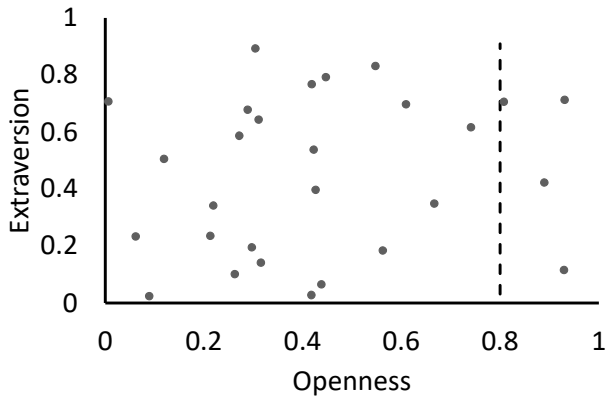

Fig. 1 An illustration of randomly distributed data naturally forming fuzzy clusters along the dimensions (axes) of the data.

Supervised feature extraction via a predictive recurrent neural network, however, yielded more constructive results. Using the established elbow-method, we determined that a network with three internal LSTM nodes was the point of diminishing returns; more than three nodes did not significantly increase the predictive accuracy, while fewer nodes substantially reduced the accuracy. These three nodes represented the extracted features and since LSTM nodes have memory the features could be visualized as trajectories in time. In Fig. 2 we visualize all combinations of the two-dimensional projections of the three-dimensional state space. We found that the extracted features for each customer followed trajectories with low average change of direction.

Furthermore, the trajectories corresponding to dominant personality traits formed clusters. We also observed a hierarchy of sub-clusters for lesser personality traits; as we zoomed into a cluster for a personality trait, we recursively found sub-clusters which corresponded to lesser personality traits. In other words, the existing hierarchy of the relative strength of the personality traits was reflected in a hierarchy of clusters of spending trajectories. This hierarchy of trajectories could be used for micro-segmentation to personalize financial recommendations. Additionally, these clusters were stable in time, as each trajectory remained in the same micro-cluster for the observed six-year time period. In this study we merely observed the presence of the clusters, but in future work we intend to apply more formal trajectory clustering techniques, which typically have a complexity of $O(n)$, as described in [21].

The formation of these trajectories in time is an interesting observation, since no such trajectories were present in raw input data – transaction classes aggregated in time. Interestingly, we found similar trajectories in the state space of a recurrent *autoencoder* as in the *predictor*, but with no clustering. A possible reason for the lack of trajectories with low change in direction in the raw input data is the natural inconsistency of spending; events naturally occur in people's lives that suddenly and temporarily require a different spending pattern, e.g., large purchases such as cars or irregular expenditure such as medical bills or household repairs. However, it seems that recurrent neural networks are able to 'smooth' these naturally inconsistent data. We hypothesize that the recurrent neural network managed to learn temporal trajectories from the input (as observed in the autoencoder) and clustering from the output. Interestingly, features extracted from a feed-forward neural network behaved differently: though they formed clusters along the dominant personality trait, no sub-clustering was observed. Naturally, without a time element, there were also no trajectories and no temporal stabilization in feed-forward networks.

### B. Transfer Learning Case Study

To test the efficacy of our extracted features, we benchmarked them against randomly initialized models of identical architectures. Table 1 shows the predictive performance of our extracted features on customer liquidity index, while Table 2 shows the performance when predicting loan default rate. In both cases, our extracted features performed at least as well as the randomly initialized models, but with far fewer trainable parameters. Having fewer trainable parameters has several benefits, including reduced training time and smaller dataset requirements. We also noticed that the confidence intervals were typically smaller for the transfer learning cases, suggesting improved precision.

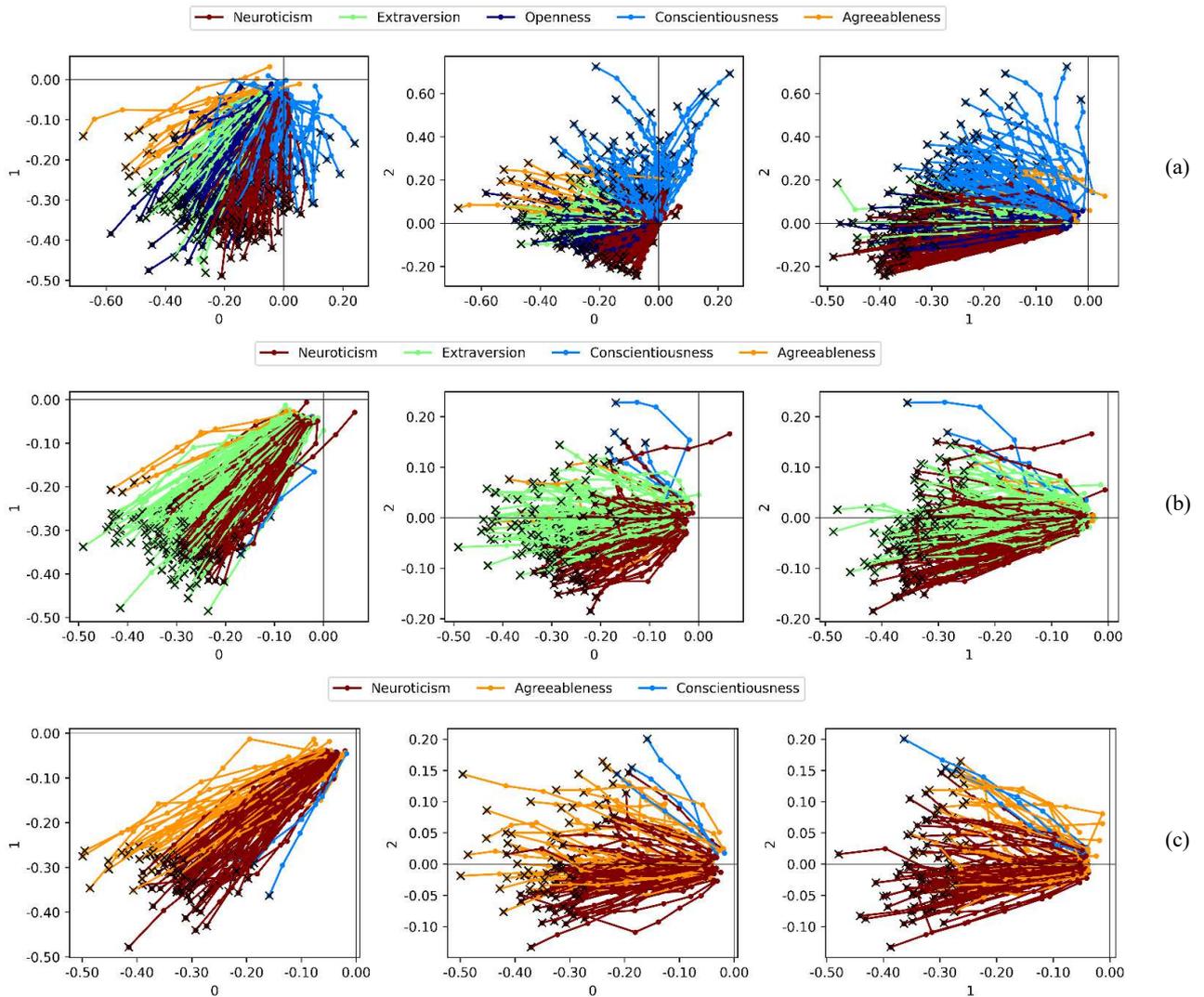

Fig. 2 Hierarchical clustering of customers' spending personalities illustrated in three parts: (a) through (c). We show the two-dimensional projections of the trajectories from the state space of a recurrent neural network where each axis represents the activation of a single node. Each trajectory represents the annual aggregated spending of a single customer for a six-year period. Part (a) shows the clustering of customers' trajectories by their dominant personality trait, while parts (b) and (c) drill down to show sub-clusters of trajectories corresponding to the second and third most dominant traits, respectively. (b) shows the sub-clusters for the parent cluster "Openness", while (c) drills down into the sub-cluster "Extraversion" from (b).

TABLE 1 A COMPARISON OF THE MEAN SQUARE ERROR VALIDATION LOSSES WHEN PREDICTING *LIQUIDITY INDEX* ON A LIMITED DATASET USING TRANSFER LEARNING VERSUS RANDOMLY INITIALIZED WEIGHTS. IN EACH CASE, THE TRANSFER LEARNING MODEL DID AT LEAST AS WELL AS A RANDOMLY INITIALIZED MODEL BUT HAD FAR FEWER TRAINABLE WEIGHTS.

| Neural network Type | Weight initiali-zation | Total weights | Trainable weights | MSE loss | 95% confidence interval |
|---|---|---|---|---|---|
| Recurrent | Random | 1637 | 1637 | 0.81 | 0.012 |
| Recurrent | Transfer learning | 1637 | 5 | 0.81 | 0.016 |
| Feed-forward | Random | 506 | 506 | 0.81 | 0.048 |
| Feed-forward | Transfer learning | 506 | 6 | 0.81 | 0.035 |

TABLE 2 A COMPARISON OF THE MEAN SQUARE ERROR VALIDATION LOSSES WHEN PREDICTING DEFAULT RATE ON A LIMITED DATASET USING TRANSFER LEARNING VERSUS RANDOMLY INITIALIZED WEIGHTS. IN EACH CASE, THE TRANSFER LEARNING MODEL DID AT LEAST AS WELL AS A RANDOMLY INITIALIZED MODEL BUT HAD FAR FEWER TRAINABLE WEIGHTS.

| Neural network Type | Weight initiali-zation | Total weights | Trainable weights | MSE loss | 95% confidence interval |
|---|---|---|---|---|---|
| Recurrent | Random | 1637 | 1637 | 11.4 | 0.048 |
| Recurrent | Transfer learning | 1637 | 5 | 11.4 | 0.006 |
| Feed-forward | Random | 506 | 506 | 6.7 | 0.125 |
| Feed-forward | Transfer learning | 506 | 6 | 6.4 | 0.002 |

## V. Conclusions and Directions for Future Work

In this paper, we introduce a novel approach for customer micro-segmentation by extracting features from customers' financial transactions using recurrent neural networks. We used published coefficients to calculate customers' personalities which we used for feature extraction. We found that by using recurrent neural networks we were able to introduce an element of time to the transactions, which stabilized the extracted features and facilitated micro-segmentation. The features followed trajectories with low average changes in direction in the extracted feature space – meaning the customers remained within their micro-segments for the observed time frame – which was not the case for their spending data or their calculated personalities. These trajectories could be recursively sub-clustered according to successive dominance of customers' personality traits, leading to a hierarchy of sub-clusters. This hierarchy of customer spending trajectories is important because could be used for micro-segmentation which might facilitate personalized financial services.

We demonstrated the efficacy of our extracted features in a transfer learning case study predicting both loan default rate and customer liquidity index. We benchmarked our transfer learning models against randomly initialized models of identical architectures. We found that our extracted features performed at least as well as randomly initialized models but required far fewer trainable parameters. Fewer trainable parameters pose several benefits in a neural network, including faster training times and smaller dataset requirements.

In future work, we want to test our hypothesis that the extracted feature trajectories are robust with respect to the window of aggregation. This will be an improvement on the clustering behavior observed in spending personality, which is erratic for shorter time windows. Having such stable micro-segments will allow the development of personalized financial services, such as budgeting and savings advice. Each customer trajectory places that customer on a 'ski slope' in the state space of the recurrent neural network, indicating a pattern in spending personality. Personality is expected to play a significant role, as it has been shown that happiness is increased when spending fits personality [9]. We will also apply formal trajectory clustering methods as described in [21] and inspect the impact of noisy data and occurrence of outliers. Finally, we intend to provide a formal explanation for our extracted features and an interpretation of our model.

ACKNOWLEDGMENT

We thank Joe Gladstone for insightful conversations about personality and spending and Perry McPartland for proofreading the first draft of the manuscript.

References

[1] A. Fernández, "Artificial intelligence in financial services," The Bank of Spain, 2019.

[2] K. Simonyan and A. Zisserman, "Very deep convolutional networks for large-scale image recognition," in *3rd International Conference on Learning Representations*, San Diego, USA, 2015.

[3] J. Xiao, R. Wang, G. Teng and Y. Hu, "A transfer learning based classifier ensemble model for customer credit scoring," in *Seventh International Joint Conference on Computational Sciences and Optimization*, Beijing, China, 2014.

[4] B. Zhu, J. Xiao and C. He, "A balanced transfer learning model for customer churn prediction," in *Proceedings of the Eighth International Conference on Management Science and Engineering Management*, Berlin, Germany, 2014.

[5] E. T. Apeh, B. Gabrys and A. Schierz, "Customer profile classification using transactional data," in *Third World Congress on Nature and Biologically Inspired Computing*, Salamanca, Spain, 2011.

[6] W. R. Smith, "Product differentiation and market segmentation as alternative marketing strategies," *The Journal of Marketing,* vol. 21, no. 1, pp. 3-8, 1956.

[7] S. Barocas and A. Selbst, "Big data's disparate impact," *California Law Review,* vol. 104, no. 671, pp. 671-732, 2016.

[8] J. J. Gladstone, S. C. Matz and A. Lemaire, "Can psychological traits be inferred from spending? Evidence from transaction data," *Psychological Science,* vol. 30, no. 7, pp. 1-10, 2019.

[9] S. C. Matz, J. J. Gladstone and D. Stillwell, "Money buys happiness when spending fits our personality," *Psychological Science,* vol. 27, no. 5, pp. 715-725, 2016.

[10] B. Rammstedt and O. P. John, "Measuring personality in one minute or less: A 10-item short version of the Big Five Inventory in English and German," *Journal of Research in Personality,* vol. 41, no. 1, pp. 203-212, 2007.

[11] E. K. Nyhus and P. Webley, "The role of personality in household saving and borrowing behaviour," *European Journal of Personality,* vol. 15, no. 1, pp. 85-103, 2001.

[12] L. Mangiavacchi, L. Piccoli and C. Rapallini, "Personality traits and household consumption choices (in press)," *The B.E. Journal of Economic Analysis & Policy,* 2020.

[13] S. Mousaeirad, "Intelligent vector-based customer segmentation in the banking industry," *arXiv:2012.11876v1,* pp. 1-41, 2020.

[14] C. Maree, J. E. Modal and C. W. Omlin, "Towards responsible AI for financial transactions," in *IEEE Symposium Series on Computational Intelligence (SSCI)*, Canberra, Australia, 2020.

[15] J. Shlens, "A tutorial on principle component analysis," *arXiv:1404.1100v1,* pp. 1-12, 2014.

[16] M. A. Kramer, "Nonlinear principal component analysis using autoassociative neural networks," *American Institute of Chemical Engineers (AIChE),* vol. 37, no. 2, pp. 233-243, 1991.

[17] S. Gu, B. Kelly and D. Xiu, "Autoencoder asset pricing models," *Journal of Econometrics,* vol. 222, no. 1, pp. 429-450, 2021.

[18] K. Greff, R. K. Srivastava, J. Koutnik, B. R. Steunebrink and J. Schmidhuber, "LSTM: A search space odyssey," *IEEE transactions on neural networks and learning systems,* vol. 28, no. 10, pp. 2222-2232, 2015.

[19] C. W. Omlin and C. L. Giles, "Extraction of rules from discrete-time recurrent neural networks," *Neural Networks,* vol. 9, no. 1, pp. 41-52, 1996.

[20] C. W. Omlin and C. L. Giles, "Training second order recurrent neural networks using hints," in *Proceedings of the Ninth International Conference on Machine Learning*, San Mateo, USA, 1992.

[21] J. Bian, D. Tian, Y. Tang and D. Tao, "A survey on trajectory clustering analysis," *arXiv,* vol. 1802.06971, pp. 1-40, 2018.